\documentclass[11pt]{article}
\usepackage{coling2018}
\usepackage{times}
\usepackage{url}
\usepackage{latexsym}
\usepackage{xspace}

\usepackage{tabularx}
\usepackage{multirow}
\usepackage{subfiles}
\usepackage{xargs}
\usepackage[pdftex,dvipsnames,table]{xcolor}
\usepackage[colorinlistoftodos,prependcaption,textsize=small]{todonotes}
\usepackage{booktabs}
\usepackage{enumitem}
\usepackage{caption}
\usepackage{bm} 
\newcommandx{\unsure}[2][1=]{\todo[linecolor=red,backgroundcolor=red!25,bordercolor=red,#1]{#2}}
\newcommandx{\change}[2][1=]{\todo[linecolor=blue,backgroundcolor=blue!25,bordercolor=blue,#1]{#2}}
\newcommandx{\info}[2][1=]{\todo[linecolor=OliveGreen,backgroundcolor=OliveGreen!25,bordercolor=OliveGreen,#1]{#2}}
\newcommandx{\improvement}[2][1=]{\todo[linecolor=Plum,backgroundcolor=Plum!25,bordercolor=Plum,#1]{#2}}
\newcommandx{\thiswillnotshow}[2][1=]{\todo[disable,#1]{#2}}

\newcommand{\datasetname}{\textsc{smhd}\xspace}
\newcommand{\datasetnamerc}{\textsc{smhd-rc}\xspace}
\newcommand{\rsdd}{\textsc{rsdd}\xspace}
\newcommand{\citet}[1]{\newcite{#1}}

\newcommand{\lcb}{{\tt {\char '173}}}   

\renewcommand\footnotemark{}

\title{SMHD: A Large-Scale Resource for Exploring Online Language Usage for Multiple Mental Health Conditions}

\author{Arman Cohan$^1$ *\thanks{* Equal contribution.} \quad\quad Bart Desmet$^{1,2}$ * \quad\quad Andrew Yates$^{1,3}$ *\\
  \textbf{Luca Soldaini$^1$ ~\quad Sean MacAvaney$^1$ ~\quad Nazli Goharian$^1$}\\
  \vspace{3mm}
  \begin{tabular}{*{2}{>{\centering}p{.49\textwidth}}}
\tabularnewline
$^1$IR Lab, Georgetown University, US   & $^2$LT3, Ghent University, BE
\tabularnewline
{\small \tt \lcb firstname\rcb@ir.cs.georgetown.edu} & {\small \tt bart.desmet@ugent.be}
\tabularnewline
\tabularnewline
\multicolumn{2}{c}{$^3$Max Planck Institute for Informatics, DE}
\tabularnewline
\multicolumn{2}{c}{\small \tt ayates@mpi-inf.mpg.de}
\tabularnewline
\end{tabular}
}

\date{}

\begin{document}
\maketitle
\begin{abstract}
Mental health is a significant and growing public health concern.
As language usage can be leveraged to obtain crucial insights into mental health conditions, there is a need for large-scale, labeled, mental health-related datasets of users who have been diagnosed with one or more of such conditions.
In this paper, we investigate the creation of high-precision patterns to identify self-reported diagnoses of nine different mental health conditions, and obtain high-quality labeled data without the need for manual labelling. We introduce the \datasetname (Self-reported Mental Health Diagnoses) dataset and make it available. \datasetname is a novel large dataset of social media posts from users with one or multiple mental health conditions along with matched control users. We examine distinctions in users' language, as measured by linguistic and psychological variables. We further explore text classification methods to identify individuals with mental conditions through their language.

\end{abstract}

%
%
\blfootnote{
    %
    %
    %
    %
    %

    \hspace{-0.65cm}  
    This work is licensed under a Creative Commons
    Attribution 4.0 International License.
    License details:
    \url{http://creativecommons.org/licenses/by/4.0/}
}

\section{Introduction}
\label{sec:intro}

Mental health is a significant challenge in healthcare. Mental disorders have the potential to tremendously affect the quality of life and wellness of individuals in society \cite{strine2008depression,mowery2017feature}.
Social media have become an increasingly important source of data related to mental health conditions \cite{cohan2017triaging,mowery2017understanding,coppersmith2017scalable,yates2017depression}, as it is now a prominent platform for individuals to engage in daily discussions, share information, seek advice and simply communicate with peers that have shared interests. In addition to its ubiquity and ease of access, the possibility to disclose mental health matters anonymously or pseudo-anonymously further drives users to online self-disclosure.

At the same time, the close connection between language and mental health makes social media an invaluable resource of mental health-related data. Lack of data has been one of the key limitations to understanding and addressing the challenges the domain is facing \cite{coppersmith2014quantifying}. Data from social media can not only be used to potentially provide clinical help to users in need, but also to broaden our understanding of the various mental health conditions. Social media analysis has already been proven valuable for identifying depression \cite{coppersmith2014quantifying,yates2017depression}, suicide ideation \cite{cohan2017triaging,de2017language,kshirsagar2017detecting,desmet2018suicide}, and other conditions such as schizophrenia \cite{mitchell2015quantifying}. While social media data is abundantly available, the amount of labeled data for studying mental health conditions is limited. This is due to the high cost of annotation and the difficulty of access to experts.

Prior research has investigated self-disclosure as a means of obtaining labeled data from social media. \citet{de2013major} used it to identify new mothers and track post-partum changes in emotions. \citet{coppersmith2014quantifying} specifically focused on self-reports of mental health diagnoses. In particular, \citet{coppersmith2015adhd} constructed a dataset of various mental health conditions using Twitter statements. Finally, \citet{yates2017depression} introduced a large dataset of depressed users obtained from Reddit.

We extend the previous efforts on addressing the lack of large-scale mental health-related language data. Particularly, we propose improved data collection methods through which we can obtain high-quality large-scale datasets of labeled diagnosed conditions paired with appropriate control users. Consequently, we introduce \datasetname (Self-reported Mental Health Diagnoses), a large dataset of diverse mental health conditions that can provide further insight into the mental health-related language. We leverage self-reported diagnosis statements where a user declares to have been diagnosed with a mental health condition such as depression. Examples of self-reported diagnoses are shown in Figure \ref{fig:example-self-reported}. Our dataset can be used not only to develop methods for better identifying mental health conditions through natural language, but also allows us to investigate the characteristics of language usage within each condition. We hope the availability of this new resource will foster further research into these problems and enhance reproducibility of suggested approaches.

\begin{table}[t]
        \centering
        \renewcommand{\arraystretch}{0.8}
        \setlength{\tabcolsep}{3pt}
        \begin{tabular}{@{}lrr@{}}
      \toprule
      \multirow{2}{*}{\textbf{Condition}}           & \multirow{2}{*}{\shortstack[r]{Twitter, (Copper-\\smith et al, 2015)}}    & \multirow{2}{*}{\shortstack[r]{Reddit, \\\datasetname (ours)}}    \\
      { }&{ }& \\
      \midrule
      ADHD                & 102                        & 10,098    \\
      Anxiety             & 216                        & 8,783    \\
      Autism              & n/a                        & 2,911    \\
      Bipolar             & 188                        & 6,434    \\
      Borderline          & 101                        & n/a    \\
      Depression          & 393                        & 14,139    \\
      Eating              & 238                        & 598    \\
      OCD                 & 100                        & 2,336    \\
      PTSD                & 403                        & 2,894    \\
      Schizophrenia       & 172                        & 1,331    \\
      Seasonal Affective  & 100                        & n/a    \\
      \bottomrule
        \end{tabular}
        \captionof{table}{Comparison between the number of self-reported diagnosed users per condition in the dataset of~\citet{coppersmith2015adhd} and ours (\datasetname).}
        \label{tab:user-counts}
\end{table}
\begin{table}[t]	
        \centering
      \begin{tabular}{@{}l@{}}
        \toprule
        \textit{I was officially diagnosed with ADHD last year.} \\
        \textit{I have a diagnosed history of PTSD.} \\
        \textit{my dr just diagnosed me as schizo.}\\
        \bottomrule
      \end{tabular}
      \captionof{figure}{Examples of self-reported diagnoses statements.}\label{fig:example-self-reported}
\end{table}

Our work has the following significant distinctions compared to existing social media datasets related to mental health. Previous work has studied self-reported diagnosis posts in Twitter \cite{coppersmith2015adhd}, where the post length is limited to 140 characters.\footnote{This limitation was doubled to 280 characters in late 2017.} This makes the Twitter language use rather different from real life discussions. Instead, we use data from Reddit, an interactive discussion-centric forum without any length constraints. Our dataset contains up to two orders of magnitude more diagnosed individuals for each condition than the Twitter dataset by \citet{coppersmith2015adhd}, making it suitable for exploring more recent data-driven learning methods (see Table \ref{tab:user-counts}). We choose our control users in a systematic way that makes classification experiments on the dataset realistic.

We normalize language usage between the users: by removing specific mental health signals and discussions, we focus on patterns of language in normal (general) discussions.
 While our dataset creation method is close to \citet{yates2017depression}, we extend theirs by investigating multiple high-precision matching patterns to identify self-reported diagnoses for a range of conditions. Part of our patterns are obtained through synonym discovery. Considering relevant synonyms from reliable sources increases the variety of the diagnosed users and linguistic nuances. We also explore nine common mental health conditions while \citet{yates2017depression} focus only on depression. We explore classification methods for identifying mental health conditions through social media language and provide detailed analysis that helps us understand the differences in language usage between conditions, and between diagnosed users and controls.

Our contributions are as follows: \textit{(i)} We investigate the creation of high-precision matching patterns to identify self-reported diagnoses of nine different mental health conditions.
\textit{(ii)} We introduce a large-scale dataset of nine mental health conditions that has significant extensions to existing datasets and we make our data publicly available. Our dataset includes users who might suffer from more than one condition, thus allowing language study of interacting mental conditions.
\textit{(iii)} We investigate language characteristics of each mental health group.
\textit{(iv)} We explore classification methods for detecting users with various mental health conditions.


%

\section{Related work}
\label{sec:related}

Social media offers a considerable amount of accessible common language, attracting the attention of those who study the language of individuals with mental health conditions. Twitter is a natural source, being a popular platform that enables users to share short messages publicly. Early work used crowdsourcing to identify Twitter users who report a depression diagnosis in a survey, and proposed features that are able to identify depressed users prior to the onset of depression~\cite{Choudhury2013PredictingDV}. Others found that these characteristics hold in both English and Japanese tweets~\cite{Tsugawa2015RecognizingDF}, indicating similar cross-cultural tendencies.

Due to the cost and bias introduced by relying on surveys, work shifted to identifying mental health conditions by examining the content shared by social media users.
\citet{coppersmith2014quantifying} identified approximately 1,200 Twitter users with 4 mental health conditions (bipolar, depression, PTSD, SAD) using diagnosis statements found in tweets (e.g., ``I was diagnosed with depression''). Following this work, detailed studies were conducted on users experiencing PTSD~\cite{Coppersmith2014MeasuringPT}, and schizophrenia~\cite{mitchell2015quantifying,Ernala2017LinguisticMI}.
The shared task at the 2nd Computational Linguistics and Clinical Psychology Workshop (CLPsych 2015) focused on identifying depression and PTSD users on Twitter~\cite{Coppersmith2015CLPsych2S}. This included a set of approximately 1,800 Twitter users with self-identified diagnoses. Leading submissions to CLPsych 2015 relied on the LIWC lexicon~\cite{pennebaker2015development}, topic modeling, manual lexicons, and other domain-dependent features~\cite{Resnik2015TheUO,PreotiucPietro2015TheRO}. \citet{coppersmith2015adhd} expands the research on Twitter to eleven self-identified mental health conditions. \cite{Benton2017MultitaskLF} uses this dataset (and others) with a neural multi-task learning approach to identify language characteristics. \citet{mowery2016towards} investigates specific symptoms of depression in tweets, including depressed mood, disturbed sleep, and loss of energy.

While the abundant short texts of Twitter can provide some insight into language characteristics of those with mental health conditions, long-form content can provide additional linguistic insights.
Some have investigated the language of users of an online crisis forum to identify posts of users who are at highest risk to allow for faster intervention~\cite{Milne2016CLPsych2S,cohan2017triaging}. \citet{Losada2016ATC} applied the self-reported diagnosis strategy to identify approximately 150 Reddit users who suffer from depression, and paired them with 750 control users. \citet{yates2017depression} also used self-reported diagnoses to identify clinically depressed users, but applied it to a larger set of Reddit, yielding the Reddit Self-reported Depression Diagnosis (\rsdd) dataset of over 9,000 users with depression and over 100,000 control users (using an improved user control identification technique). The corpus was also used to study the temporal aspects of self-reported diagnoses \cite{macavaney2018rsdd}.

Others have used data sources beyond social media to examine the language of people with mental health conditions. \citet{Resnik2013UsingTM} uses topic models to predict depression and neuroticism based on student-written essays, finding clear clusters of words when students are asked to write about their feelings in a stream-of-consciousness setting. \citet{Althoff2016LargescaleAO} uses text message conversations from a mental health crisis center to improve counseling techniques.

This work addresses important limitations of previous efforts. Similar to \rsdd~\cite{yates2017depression}, we build our corpus from Reddit using self-reported diagnoses. This results in a large amount of long-form post content that is not constrained by a character limit. Furthermore, because there are no character limits (as exist for Twitter), the mode of language is more typical of general writing. Unlike \citet{yates2017depression}, we investigate extended self-diagnoses matching patterns derived from mental health-related synonyms. We also focus on nine mental health conditions (rather than just a single condition). This results in a collection that can be used to compare and contrast the language characteristics of each condition.

\section{Data}
\label{sec:data}


In this section we describe the construction and characteristics of the Self-reported Mental Health Diagnoses (\datasetname) dataset.
The studied conditions correspond to branches in the DSM-5~\cite{apa2013diagnostic}, an authoritative taxonomy for psychiatric diagnoses.
Six conditions are top-level DSM-5 disorders: schizophrenia spectrum disorders (\textit{schizophrenia}), bipolar disorders (\textit{bipolar}), depressive disorders (\textit{depression}), anxiety disorders (\textit{anxiety}), obsessive-compulsive disorders (\textit{ocd}) and feeding and eating disorders (\textit{eating}).
The three other conditions are one rank lower: post-traumatic stress disorder (\textit{ptsd}) is classified under trauma- and stress-related disorders, and autism spectrum disorders (\textit{autism}) and attention-deficit/hyperactivity disorder (\textit{adhd}) under neurodevelopmental disorders.
We use these lower-rank conditions to provide more definition when they are clearly distinguishable from sibling disorders.

\subsection{Dataset construction}
The \datasetname dataset was created by using high precision patterns to identify Reddit users who claimed to have been diagnosed
with a mental health condition (\textit{diagnosed users})
and using exclusion criteria to match these diagnosed users with control users who are unlikely
to have one of the mental health conditions studied (\textit{control users}).
\datasetname consists of user labels indicating the mental health condition(s) associated with each user
and all Reddit posts made by each user between January 2006 and December 2017 (inclusive).
Users and posts were extracted from a publicly available Reddit corpus\footnote{\url{https://files.pushshift.io/reddit/}}.
This approach is based on the method used to create the Reddit Self-reported Depression Diagnosis (\rsdd)
dataset~\cite{yates2017depression}. \datasetname expands on \rsdd by incorporating synonyms in matching patterns and including diagnoses for eight new conditions in addition to depression.\footnote{Patterns are available from \url{http://ir.cs.georgetown.edu/data/smhd/}.}

\textbf{Diagnosed users} were identified using high precision diagnosis patterns in a manner similar to that
used in prior work that studied depression on Reddit~\cite{yates2017depression};
we describe these patterns in more detail in section~\ref{subsec:patterns}.
After identifying candidate diagnosed users who matched a diagnosis pattern (see Figure~\ref{fig:example-self-reported}),
we removed any candidates who \textit{(1)} matched a negative diagnosis pattern\footnote{e.g., \textit{I was never clinically diagnosed.}}
or \textit{(2)} had fewer than 50 posts talking about topics other than mental health (\textit{mental health posts}).
This is done to ensure enough data remains for a diagnosed user after removing their mental health-related content.
After removing these candidate diagnosed users, 36948 diagnosed users remain.

\textbf{Mental health posts} were defined as posts that were either made to a subreddit (i.e., a subforum devoted to a specific topic) related to mental health or that
included language related to mental health, such as the name of a condition (e.g., \textit{OCD}) and general terms
like \textit{diagnosis}, \textit{mental illness}, or \textit{suffering from}.
We constructed a list of subreddits related to mental health by beginning with lists from prior work studying depression on
Reddit~\cite{Pavalanathan:2015:IMM:2740908.2743049,yates2017depression} and expanding them to include discussion and support subreddits for each of the other mental health conditions.
All mental health posts are removed for diagnosed users and control users alike. Classification therefore happens on the set of posts that do not contain any of the mental health terms, and that have not been posted in any of the mental health-related subreddits. Our methodology does not guarantee, however, that all potentially relevant terms or subreddits have been excluded.

\begin{figure}[t]
    \centering
    \small
	\begin{minipage}{0.55\textwidth}
        \centering
        \small
        \includegraphics[width=0.8\textwidth]{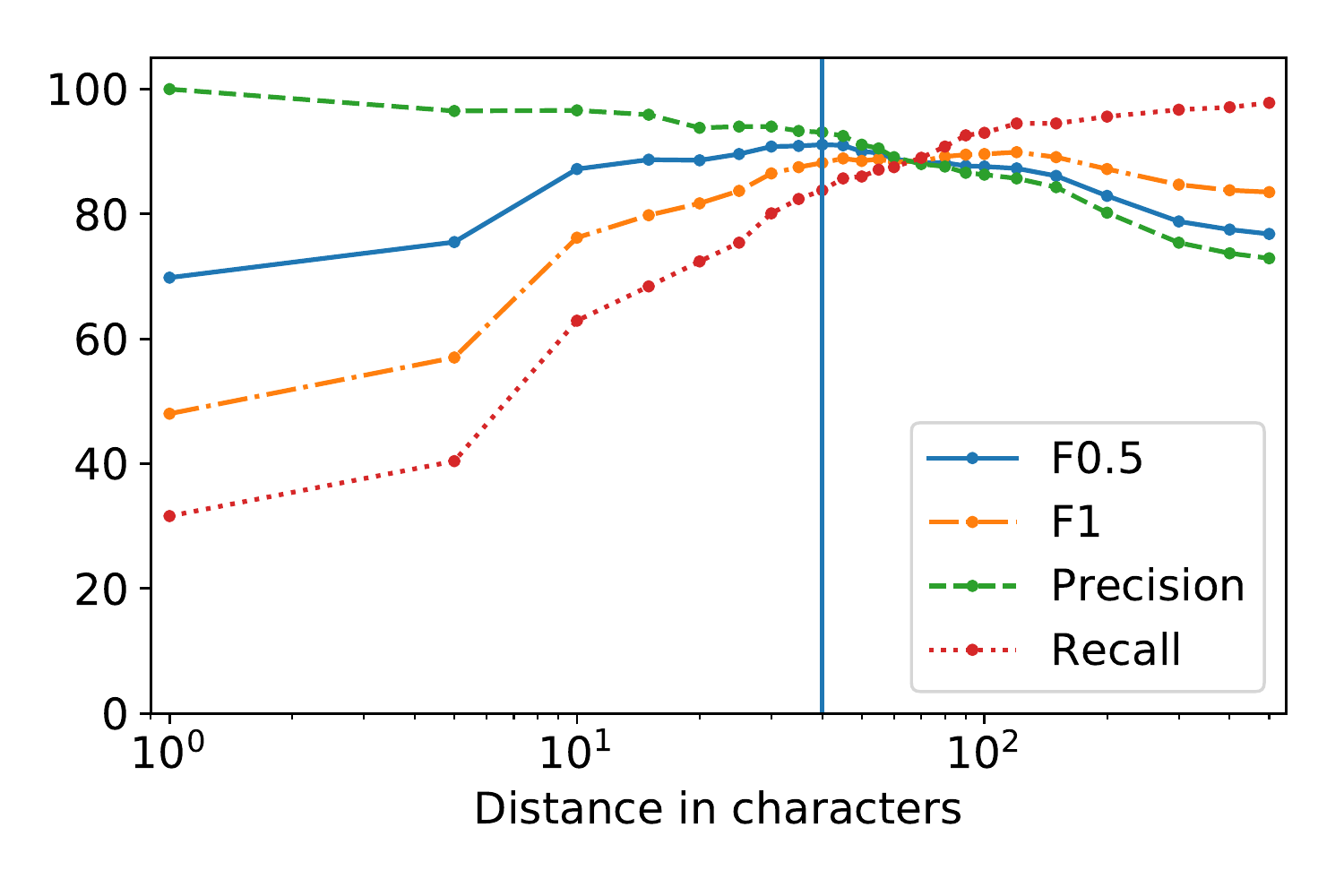}
        \caption{Precision, recall, F1 and F0.5 of the condition diagnosis patterns as a function of the maximum allowable distance (in characters) between a diagnosis and a condition keyword. Chosen threshold indicated at 40 characters.}
        \label{fig:maxmindist}
    \end{minipage}\hfill
    \begin{minipage}{0.40\textwidth}
        \centering
        \small
        \includegraphics[width=1\textwidth]{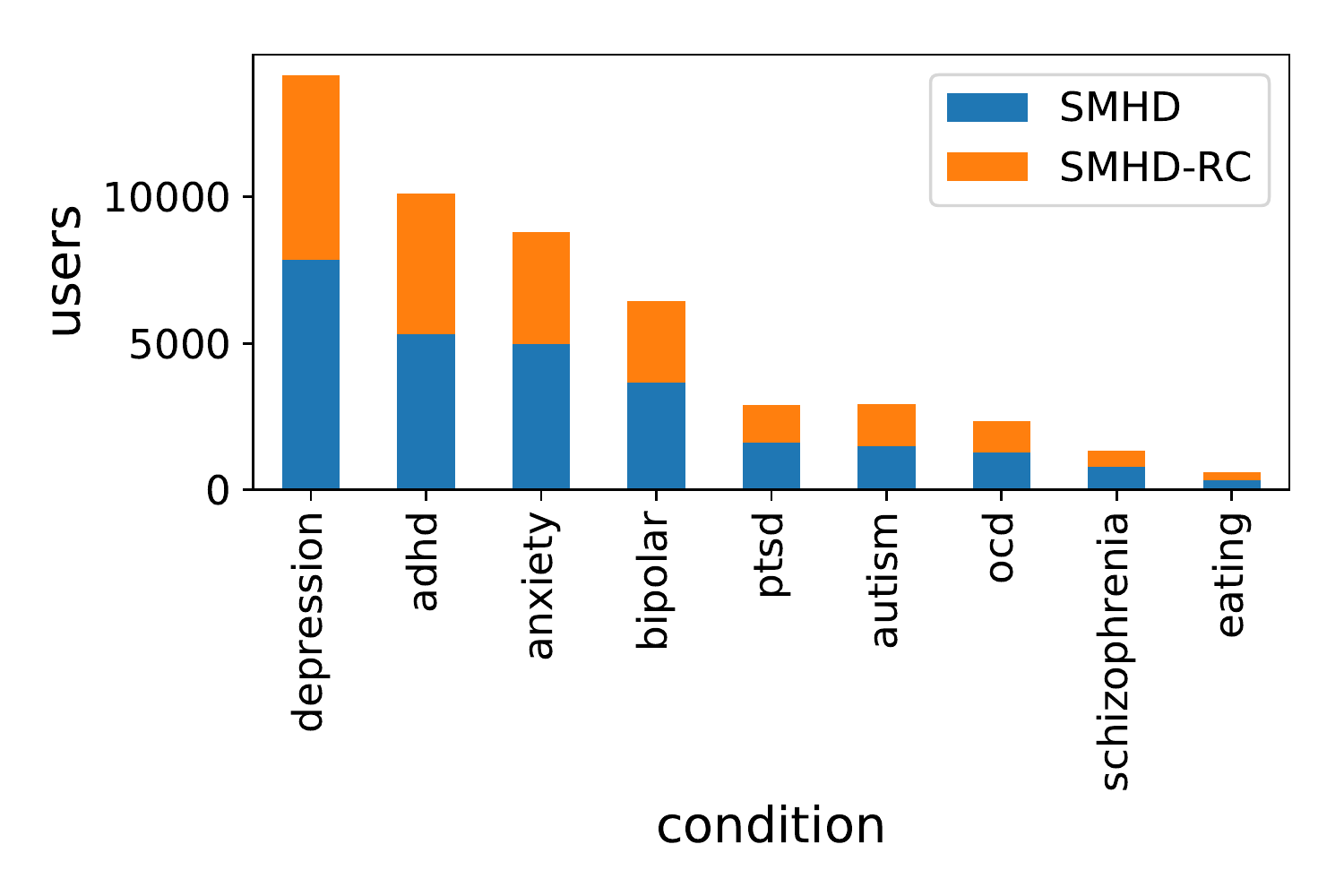}
        \caption{ Number of users per condition for \datasetname (in blue) and \datasetnamerc (in orange).}
        \label{fig:conds}
    \end{minipage}
    \vfill
    \begin{minipage}{0.45\textwidth}
        \centering
        \small
        \includegraphics[width=1\textwidth]{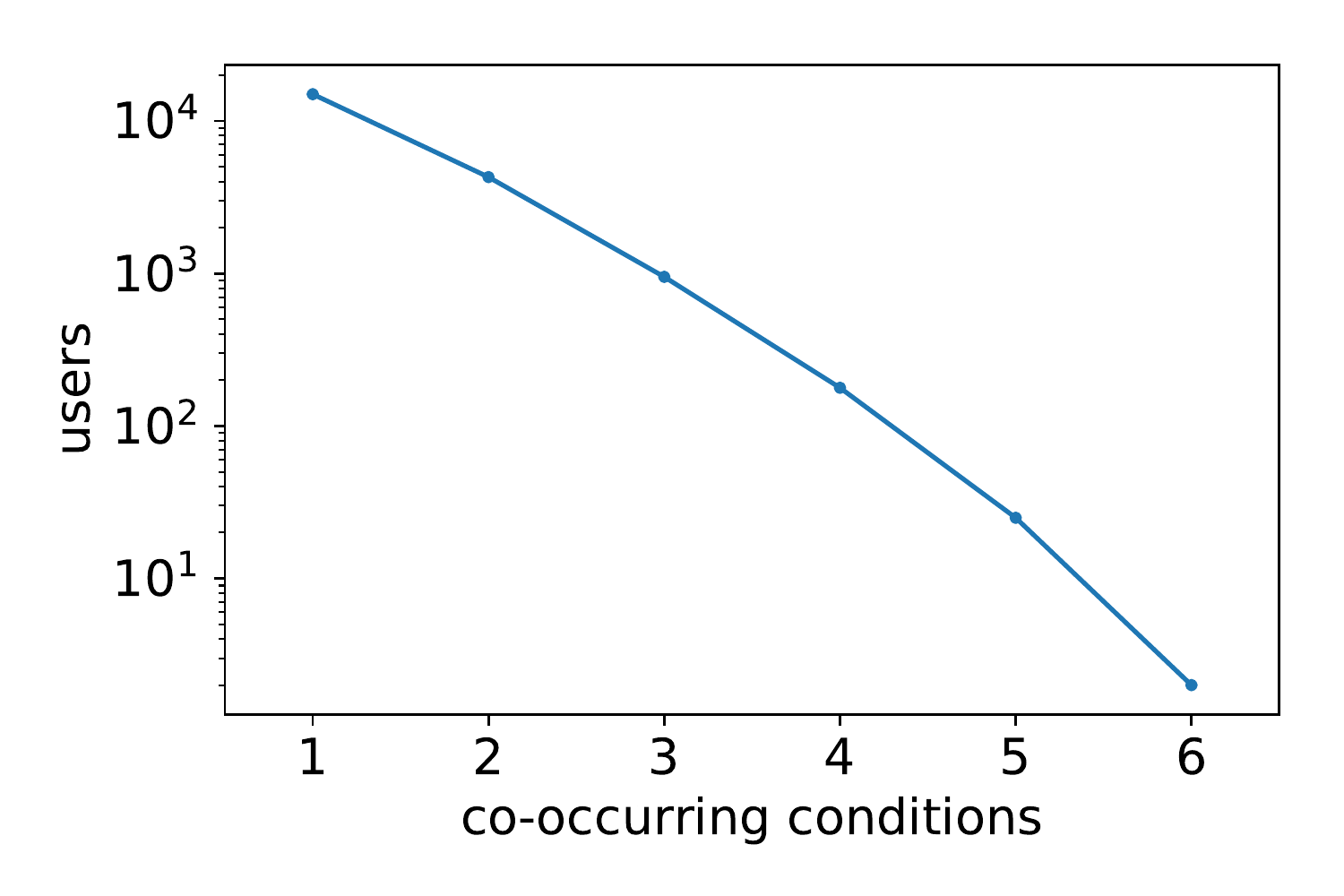}
        \caption{ Number of users with a single or multiple co-occurring conditions.}
        \label{fig:concom}
    \end{minipage}\hfill
    \begin{minipage}{0.45\textwidth}
        \centering
        \small
        \includegraphics[width=1.0\textwidth]{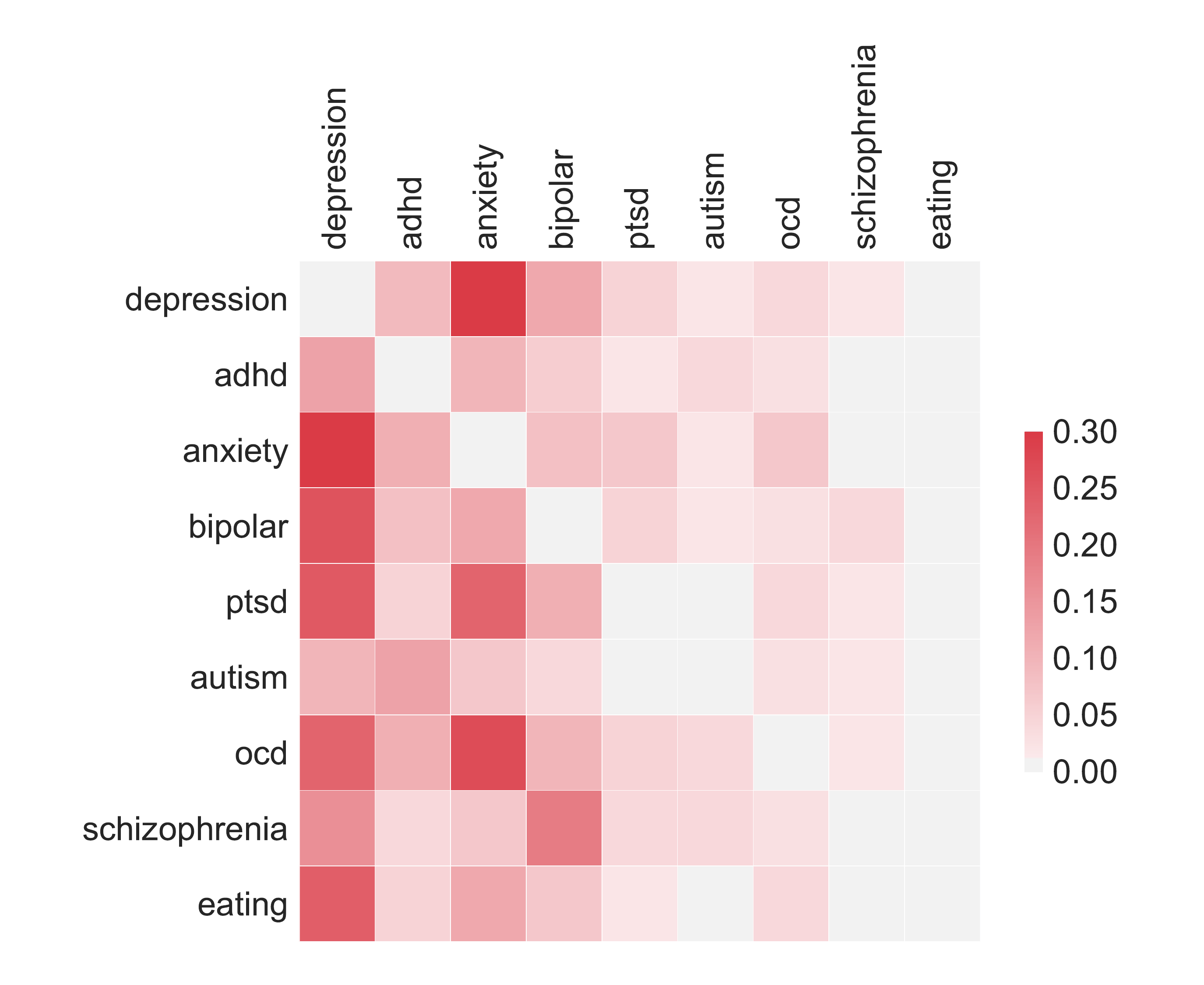}
        \caption{ Concomitance of condition diagnoses. Rows contain relative co-occurences of the row condition with other conditions.}
        \label{fig:concomitance}
    \end{minipage}
\end{figure}

\textbf{Control users} were chosen from a pool of candidate control users based on their similarity with the diagnosed users
as measured by their number of posts and the subreddits they posted in. This is done to prevent biases between the control and diagnosed groups in the dataset and prevent making the task of identifying such users artificially easy. In more detail, every Reddit user who
\textit{(1)} had no mental health post\footnote{When defining language related to mental health, we consider \textit{add} to be a mental health term. However, this term is often used as a verb that does not refer to Attention Deficit Disorder. We therefore do not exclude candidate control users who have used this term. We instead remove any post containing the term \textit{add}, which matches our treatment of the diagnosed users where all mental health posts are removed.}
and \textit{(2)} had at least 50 posts became a candidate control user.
Given this pool of candidate control users, we matched each diagnosed user with 9 candidate control users (on average)
after excluding controls who had never posted in the same subreddit as the diagnosed user or who
had more than twice as many posts or less than half as many posts as the target diagnosed user.

The selection criteria for potential control users are stringent: users are removed if they do not have the required subreddit overlap or minimum post count, or if they use any of the mental health-related terms in any of their posts.
The latter is necessary because we found that the prior probability of using an exclusion term in any given subreddit (e.g. \emph{depressed}) is almost always higher for diagnosed users than for controls, even in popular subreddits on general topics (e.g. \emph{r/politics}). 
As a result, the pool of candidate controls for matching to diagnosed users is significantly diminished. Because control users are picked from the pool without replacement, we were unable to meet the target of 9 appropriate controls for some of the diagnosed users.
We release the users who have at least 9 control users as \datasetname, and use this dataset for all analyses in this paper. It contains 20,406 diagnosed users and 335,952 matched controls.
A secondary dataset containing the remaining 16,542 diagnosed users with fewer than 9 controls will be made available as \datasetnamerc (\textit{Relaxed Controls}) for studies that require additional data for analysing differences between mental health conditions, rather than between diagnosed users and controls.

\subsection{Diagnosis patterns}
\label{subsec:patterns}
To identify Reddit users who report a diagnosis for one or more mental health conditions, we developed detection patterns with a focus on high precision. The patterns consist of two components: one that matches a self-reported diagnosis, and another that maps relevant diagnosis keywords to the 9 mental health conditions. A user is included for a condition if one of the condition keywords occurs within a certain distance of the diagnosis pattern (as discussed below).

For each condition, a seed list of diagnosis keywords was collected from the corresponding DSM-5 headings.
To increase the likelihood of matching diagnostic posts, we expanded each set of medical expressions to include synonyms, common misspellings, vernacular terms and abbreviations.
Our steps mirror the ones of \citet{soldaini2017inferring}, who were also interested in identifying self-diagnosed users, albeit on query logs.
In particular, we leveraged two synonym mappings to generate alternative formulations of the disorders of interest:
\begin{itemize}[leftmargin=6pt,noitemsep]
  \item \emph{MedSyn}~\cite{yates2013adrtrace} is a laypeople-oriented synonym mapping ontology. It was generated from a subset of \emph{UMLS}\footnote{Unified Medical Language System} filtered to remove irrelevant semantic types.
  \item \emph{Behavioral}~\cite{yom2013postmarket} maps expressions commonly used by laypeople to describe their medical condition to concepts in UMLS. The synonyms were generated by first identifying the most frequent search terms Yahoo! search used to retrieve Wikipedia medical pages. Then, frequent lexical affinities \cite{carmel2002automatic} to the aforementioned were added to synonyms lists.
\end{itemize}
The resulting expansions were vetted manually to remove terms that did not describe the condition precisely (e.g., hypernyms) or that were ambiguous (e.g., ``add''), and missing items were added.

Using the diagnosis patterns without condition keywords, all self-reported diagnosis posts were collected from the Reddit dataset, and 500 posts were randomly sampled for manual annotation. False positive matches were found in 18 of those, i.e. the precision for self-diagnosis detection was 96.4\%. False positives included negations, hypotheticals (e.g., ``I'm not sure if I was diagnosed''), and diagnoses that are uncertain or wrong (``I was diagnosed with autism, then undiagnosed.''). Out of 500 diagnosis posts, 241 reported a mental health condition. The ones that were annotated as belonging to one of the 9 conditions were used to tune the condition term lists and the optimal distance between a diagnosis pattern and a condition term.
Figure~\ref{fig:maxmindist} plots the effect of this distance on precision, recall, F1 and F0.5 (which emphasizes precision). A maximum distance of 40 characters was chosen, where F0.5 score is highest. We thus achieve high precision (93\%) with good recall (84\%).
Since the optimal distance threshold was tuned on this development set, it may overfit the data and the reported scores should be considered a ceiling performance.

To validate the final diagnosis matching approach on a held-out set, 900 posts (corresponding to 9 samples of 100 diagnosis posts, one for each condition) was manually checked for false positives. We obtain high precision, with a minimum precision of 90\% for \textit{anxiety} and macro-averaged precision of 95.8\%. Most false positives are caused by terms for a condition occurring close to a diagnosis for another condition (e.g. ``My doctor diagnosed me with depression, and I also have an anxiety problem.''). While a user might also suffer from these conditions, they are not explicitly reporting a diagnosis.

\subsection{Dataset statistics}
\label{sec:stats}

Figure~\ref{fig:conds} shows the distribution of diagnosed users per condition in both datasets. Users who self-reported a diagnosis of depression, ADHD, anxiety or bipolar are most common.
Interestingly, 26.7\% of diagnosed users in the dataset reported more than one diagnosis (Figure~\ref{fig:concom}). Such concomitant conditions are not uncommon, and were also reported in the work of \citet{coppersmith2015adhd}. As can be seen in Figure~\ref{fig:concomitance}, depression co-occurs with high frequency with most of the other conditions, almost 30\% of users with depression, OCD or PTSD also suffer from anxiety, and schizophrenia is most likely to be diagnosed alongside bipolar disorder.



An important characteristic of the \datasetname dataset is its scale.
Reddit allows its users to write long-form posts, so unlike datasets collected on Twitter, a large amount of text is available for each diagnosed user.
Table~\ref{tab:lengths} gives an overview of the average and total number of posts, tokens and characters per condition, and for controls.
Control users post on average twice as many posts as diagnosed users, but these posts tend to be considerably shorter. 
Although this may be a valid signal for certain mental health conditions, it can be removed for classification experiments by truncating the length and number of posts. This is common practice for technical reasons, and truncating post length has been shown in previous work to improve classification performance \cite{yates2017depression}.

\begin{table}[t]
  \centering
  \small
  \setlength{\tabcolsep}{3pt}
  \begin{tabular}{lrrrrrr}
\toprule
\textbf{condition}		& \multicolumn{2}{c}{\textbf{posts}}	&\multicolumn{2}{c}{\textbf{tokens}}	& \textbf{characters}	\\
                        \cmidrule{2-3}                   \cmidrule(l){4-5}                     \cmidrule(l){6-6}
				& per user		& total		& per post		& total		& per post		\\
\midrule
control			& 310.0 (157.8)	& 115,669k	& 26.2 (48.3)	& 3,031.6M	& 133.9 (252.9)	\\
\midrule
depression		& 162.2 (84.2)	& 1,272k	& 45.1 (80.0)	& 57.4M		& 227.5 (406.9)	\\
adhd			& 164.7 (83.6)	& 872k		& 46.5 (82.7)	& 40.5M		& 237.5 (433.5)	\\
anxiety			& 159.7 (83.0)	& 795k		& 46.4 (83.0)	& 36.9M		& 233.9 (422.8)	\\
bipolar			& 157.6 (82.4)	& 575k		& 45.5 (86.5)	& 26.2M		& 230.6 (447.0)	\\
ptsd			& 160.7 (84.7)	& 258k		& 53.1 (114.0)	& 13.7M		& 267.8 (581.7)	\\
autism			& 168.3 (84.5)	& 248k		& 46.5 (82.3)	& 11.6M		& 237.9 (434.0)	\\
ocd				& 158.8 (81.4)	& 203k		& 46.4 (90.1)	& 9.4M		& 234.2 (459.5)	\\
schizophrenia	& 157.3 (80.5)	& 123k		& 49.2 (105.6)	& 6.1M		& 253.8 (566.6)	\\
eating			& 161.4 (81.0)	& 53k		& 46.3 (73.7)	& 2.5M		& 232.6 (372.8)	\\
\bottomrule
  \end{tabular}
  \caption{Average (Stdev.) and count of posts, tokens and characters for diagnosed and control users.}
  \label{tab:lengths}
\end{table}



\subsection{Ethics and privacy}
Even though we rely on publicly available Reddit posts in our work, mental health is a sensitive matter and measures to prevent risk to individuals in social media research should always be considered \cite{hovy2016social,vsuster2017short}. The risks associated with the data collection methods and our resulting \datasetname dataset is minimal. We refrain from publicly posting any excerpts of the data, we made no attempt to contact users, and we made no attempt to identify or link users to other social media accounts. We further replace usernames with random identifiers to prevent users' identities from being known without the use of external information. The \datasetname dataset is
available through a Data Usage Agreement (DUA)\footnote{\url{http://ir.cs.georgetown.edu/resources/}} protecting the users' privacy. In particular, the DUA specifies that no attempt should be made to publish portions of the dataset (which could result in users being identified), contact users, identify them, or link them with other user information.

\section{Analysis}
\label{sec:analysis}

To investigate the differences between the language of mental health condition groups and the control group, we categorize language of users based on measures of psycholinguistic attributes through the LIWC lexicon \cite{pennebaker2015development}. These categories include variables that characterize linguistic style as well as psychological aspects of language (e.g., cognitive attributes and affective attributes). For each user, we obtain LIWC categories based on their posts and then compare these categories across users in each mental health condition group versus the control group using Welch's t-test \cite{welch1947generalization}. We adjust p-values with Bonferroni correction. To better see the differences between the categories, we also report the Cohen's $d$  statistic \cite{cohen1988statistical}. Table \ref{tab:liwc-all} shows the results.

\begin{table}[t]
  \centering
  \scriptsize
  \setlength{\tabcolsep}{3pt}
  \renewcommand{\arraystretch}{0.8}
  \begin{tabular}{@{}lrrrrrrrrr}
  \toprule
                           \textbf{LIWC category} &                           \textbf{depression} &                                 \textbf{adhd} &                              \textbf{anxiety} &                              \textbf{bipolar} &                                 \textbf{ptsd} &                              \textbf{autism} &                                 \textbf{ocd} &                      \textbf{schizophrenia} &                               \textbf{eating} \\
  \midrule
   \textit{{Summary Language Variables}} &                                      &                                      &                                      &                                      &                                      &                                     &                                     &                                    &                                      \\
                       \hspace{3pt}Clout &  \cellcolor{blue!45}$-0.06^\ddagger$ &                                   -- &   \cellcolor{blue!45}$-0.1^\ddagger$ &                                   -- &                                   -- &                                  -- &                                  -- &                                 -- &                                   -- \\
                   \hspace{3pt}Authentic &     \cellcolor{red!45}$0.2^\ddagger$ &    \cellcolor{red!45}$0.15^\ddagger$ &    \cellcolor{red!45}$0.22^\ddagger$ &    \cellcolor{red!45}$0.18^\ddagger$ &    \cellcolor{red!45}$0.21^\ddagger$ &   \cellcolor{red!45}$0.14^\ddagger$ &   \cellcolor{red!45}$0.23^\ddagger$ &                                 -- &       \cellcolor{red!12}$0.24^\star$ \\
                         \hspace{3pt}WPS &    \cellcolor{red!45}$0.08^\ddagger$ &     \cellcolor{red!45}$0.1^\ddagger$ &    \cellcolor{red!45}$0.08^\ddagger$ &    \cellcolor{red!45}$0.06^\ddagger$ &     \cellcolor{red!45}$0.1^\ddagger$ &   \cellcolor{red!45}$0.12^\ddagger$ &    \cellcolor{red!27}$0.08^\dagger$ &                                 -- &                                   -- \\
                         \hspace{3pt}Dictionary words &    \cellcolor{red!45}$0.27^\ddagger$ &    \cellcolor{red!45}$0.22^\ddagger$ &    \cellcolor{red!45}$0.28^\ddagger$ &    \cellcolor{red!45}$0.24^\ddagger$ &    \cellcolor{red!45}$0.31^\ddagger$ &    \cellcolor{red!45}$0.2^\ddagger$ &   \cellcolor{red!45}$0.28^\ddagger$ &  \cellcolor{red!45}$0.22^\ddagger$ &     \cellcolor{red!45}$0.3^\ddagger$ \\
        \hspace{3pt}Total function words &    \cellcolor{red!45}$0.27^\ddagger$ &    \cellcolor{red!45}$0.21^\ddagger$ &    \cellcolor{red!45}$0.28^\ddagger$ &    \cellcolor{red!45}$0.24^\ddagger$ &     \cellcolor{red!45}$0.3^\ddagger$ &   \cellcolor{red!45}$0.26^\ddagger$ &   \cellcolor{red!45}$0.28^\ddagger$ &  \cellcolor{red!45}$0.23^\ddagger$ &    \cellcolor{red!45}$0.27^\ddagger$ \\
              \hspace{3pt}Total pronouns &    \cellcolor{red!45}$0.22^\ddagger$ &    \cellcolor{red!45}$0.14^\ddagger$ &    \cellcolor{red!45}$0.24^\ddagger$ &     \cellcolor{red!45}$0.2^\ddagger$ &    \cellcolor{red!45}$0.25^\ddagger$ &   \cellcolor{red!45}$0.17^\ddagger$ &   \cellcolor{red!45}$0.26^\ddagger$ &  \cellcolor{red!45}$0.18^\ddagger$ &     \cellcolor{red!27}$0.26^\dagger$ \\
           \hspace{3pt}Personal pronouns &    \cellcolor{red!45}$0.23^\ddagger$ &    \cellcolor{red!45}$0.14^\ddagger$ &    \cellcolor{red!45}$0.26^\ddagger$ &    \cellcolor{red!45}$0.21^\ddagger$ &    \cellcolor{red!45}$0.22^\ddagger$ &   \cellcolor{red!45}$0.14^\ddagger$ &   \cellcolor{red!45}$0.23^\ddagger$ &   \cellcolor{red!45}$0.2^\ddagger$ &     \cellcolor{red!27}$0.27^\dagger$ \\
           \hspace{3pt}1st pers singular &    \cellcolor{red!45}$0.23^\ddagger$ &    \cellcolor{red!45}$0.16^\ddagger$ &    \cellcolor{red!45}$0.28^\ddagger$ &    \cellcolor{red!45}$0.22^\ddagger$ &    \cellcolor{red!45}$0.23^\ddagger$ &   \cellcolor{red!45}$0.17^\ddagger$ &   \cellcolor{red!45}$0.26^\ddagger$ &   \cellcolor{red!27}$0.17^\dagger$ &     \cellcolor{red!27}$0.28^\dagger$ \\
           \hspace{3pt}3rd pers singular &    \cellcolor{red!45}$0.09^\ddagger$ &                                   -- &     \cellcolor{red!45}$0.1^\ddagger$ &       \cellcolor{red!12}$0.08^\star$ &       \cellcolor{red!12}$0.17^\star$ &                                  -- &                                  -- &                                 -- &                                   -- \\
         \hspace{3pt}Impersonal pronouns &    \cellcolor{red!45}$0.06^\ddagger$ &       \cellcolor{red!12}$0.05^\star$ &     \cellcolor{red!27}$0.07^\dagger$ &                                   -- &     \cellcolor{red!27}$0.11^\dagger$ &      \cellcolor{red!12}$0.09^\star$ &    \cellcolor{red!27}$0.13^\dagger$ &                                 -- &                                   -- \\
                \hspace{3pt}Prepositions &    \cellcolor{red!45}$0.12^\ddagger$ &    \cellcolor{red!45}$0.12^\ddagger$ &    \cellcolor{red!45}$0.12^\ddagger$ &    \cellcolor{red!45}$0.12^\ddagger$ &    \cellcolor{red!45}$0.16^\ddagger$ &   \cellcolor{red!45}$0.12^\ddagger$ &    \cellcolor{red!27}$0.11^\dagger$ &                                 -- &                                   -- \\
             \hspace{3pt}Auxiliary verbs &    \cellcolor{red!45}$0.12^\ddagger$ &     \cellcolor{red!45}$0.1^\ddagger$ &    \cellcolor{red!45}$0.14^\ddagger$ &    \cellcolor{red!45}$0.11^\ddagger$ &    \cellcolor{red!45}$0.14^\ddagger$ &   \cellcolor{red!45}$0.15^\ddagger$ &   \cellcolor{red!45}$0.13^\ddagger$ &                                 -- &                                   -- \\
              \hspace{3pt}Common Adverbs &    \cellcolor{red!45}$0.09^\ddagger$ &    \cellcolor{red!45}$0.07^\ddagger$ &    \cellcolor{red!45}$0.08^\ddagger$ &       \cellcolor{red!12}$0.06^\star$ &                                   -- &                                  -- &                                  -- &                                 -- &                                   -- \\
                \hspace{3pt}Conjunctions &    \cellcolor{red!45}$0.17^\ddagger$ &    \cellcolor{red!45}$0.14^\ddagger$ &    \cellcolor{red!45}$0.18^\ddagger$ &    \cellcolor{red!45}$0.16^\ddagger$ &    \cellcolor{red!45}$0.17^\ddagger$ &   \cellcolor{red!45}$0.13^\ddagger$ &    \cellcolor{red!27}$0.11^\dagger$ &                                 -- &     \cellcolor{red!45}$0.3^\ddagger$ \\
                  \textit{Other Grammar} &                                      &                                      &                                      &                                      &                                      &                                     &                                     &                                    &                                      \\
                \hspace{3pt}Common verbs &    \cellcolor{red!45}$0.15^\ddagger$ &    \cellcolor{red!45}$0.11^\ddagger$ &    \cellcolor{red!45}$0.17^\ddagger$ &    \cellcolor{red!45}$0.13^\ddagger$ &    \cellcolor{red!45}$0.15^\ddagger$ &   \cellcolor{red!45}$0.13^\ddagger$ &   \cellcolor{red!45}$0.19^\ddagger$ &                                 -- &                                   -- \\
                     \hspace{3pt}Numbers &   \cellcolor{blue!45}$-0.1^\ddagger$ &  \cellcolor{blue!45}$-0.09^\ddagger$ &  \cellcolor{blue!45}$-0.11^\ddagger$ &  \cellcolor{blue!45}$-0.09^\ddagger$ &  \cellcolor{blue!45}$-0.11^\ddagger$ &                                  -- &  \cellcolor{blue!45}$-0.1^\ddagger$ &                                 -- &   \cellcolor{blue!27}$-0.13^\dagger$ \\
        \textit{Psychological Variables} &                                      &                                      &                                      &                                      &                                      &                                     &                                     &                                    &                                      \\
            \hspace{3pt}Positive emotion &                                   -- &                                   -- &                                   -- &                                   -- &                                   -- &    \cellcolor{blue!12}$-0.08^\star$ &                                  -- &                                 -- &                                   -- \\
                     \hspace{3pt}Anxiety &     \cellcolor{red!27}$0.07^\dagger$ &                                   -- &       \cellcolor{red!12}$0.07^\star$ &                                   -- &                                   -- &                                  -- &                                  -- &                                 -- &                                   -- \\
            \hspace{3pt}Social processes &    \cellcolor{red!45}$0.11^\ddagger$ &    \cellcolor{red!45}$0.07^\ddagger$ &    \cellcolor{red!45}$0.11^\ddagger$ &     \cellcolor{red!45}$0.1^\ddagger$ &    \cellcolor{red!45}$0.15^\ddagger$ &                                  -- &                                  -- &                                 -- &                                   -- \\
                      \hspace{3pt}Family &    \cellcolor{red!45}$0.06^\ddagger$ &                                   -- &       \cellcolor{red!12}$0.06^\star$ &                                   -- &     \cellcolor{red!27}$0.12^\dagger$ &                                  -- &                                  -- &                                 -- &                                   -- \\
           \hspace{3pt}Female references &    \cellcolor{red!45}$0.13^\ddagger$ &    \cellcolor{red!45}$0.07^\ddagger$ &     \cellcolor{red!45}$0.1^\ddagger$ &    \cellcolor{red!45}$0.13^\ddagger$ &     \cellcolor{red!27}$0.22^\dagger$ &                                  -- &                                  -- &                                 -- &                                   -- \\
         \hspace{3pt}Cognitive processes &    \cellcolor{red!45}$0.12^\ddagger$ &    \cellcolor{red!45}$0.13^\ddagger$ &    \cellcolor{red!45}$0.14^\ddagger$ &    \cellcolor{red!45}$0.09^\ddagger$ &     \cellcolor{red!27}$0.12^\dagger$ &   \cellcolor{red!45}$0.16^\ddagger$ &    \cellcolor{red!27}$0.13^\dagger$ &                                 -- &                                   -- \\
                     \hspace{3pt}Insight &    \cellcolor{red!45}$0.09^\ddagger$ &    \cellcolor{red!45}$0.07^\ddagger$ &     \cellcolor{red!45}$0.1^\ddagger$ &    \cellcolor{red!45}$0.08^\ddagger$ &                                   -- &       \cellcolor{red!12}$0.1^\star$ &    \cellcolor{red!27}$0.17^\dagger$ &                                 -- &                                   -- \\
                 \hspace{3pt}Discrepancy &                                   -- &       \cellcolor{red!12}$0.06^\star$ &                                   -- &                                   -- &                                   -- &                                  -- &                                  -- &                                 -- &                                   -- \\
                   \hspace{3pt}Tentative &    \cellcolor{red!45}$0.07^\ddagger$ &    \cellcolor{red!45}$0.08^\ddagger$ &    \cellcolor{red!45}$0.08^\ddagger$ &       \cellcolor{red!12}$0.07^\star$ &                                   -- &                                  -- &                                  -- &                                 -- &                                   -- \\
             \hspace{3pt}Differentiation &    \cellcolor{red!45}$0.08^\ddagger$ &    \cellcolor{red!45}$0.08^\ddagger$ &     \cellcolor{red!45}$0.1^\ddagger$ &                                   -- &                                   -- &      \cellcolor{red!12}$0.09^\star$ &                                  -- &                                 -- &                                   -- \\
        \hspace{3pt}Biological processes &    \cellcolor{red!45}$0.06^\ddagger$ &                                   -- &                                   -- &                                   -- &                                   -- &                                  -- &                                  -- &                                 -- &                                   -- \\
                      \hspace{3pt}Health &    \cellcolor{red!45}$0.08^\ddagger$ &     \cellcolor{red!27}$0.07^\dagger$ &    \cellcolor{red!45}$0.08^\ddagger$ &    \cellcolor{red!45}$0.11^\ddagger$ &                                   -- &                                  -- &                                  -- &                                 -- &                                   -- \\
               \textit{Time orientation} &                                      &                                      &                                      &                                      &                                      &                                     &                                     &                                    &                                      \\
                  \hspace{3pt}Past focus &    \cellcolor{red!45}$0.08^\ddagger$ &       \cellcolor{red!12}$0.05^\star$ &    \cellcolor{red!45}$0.09^\ddagger$ &     \cellcolor{red!27}$0.08^\dagger$ &       \cellcolor{red!12}$0.09^\star$ &                                  -- &      \cellcolor{red!12}$0.11^\star$ &                                 -- &                                   -- \\
               \hspace{3pt}Present focus &    \cellcolor{red!45}$0.09^\ddagger$ &    \cellcolor{red!45}$0.06^\ddagger$ &     \cellcolor{red!45}$0.1^\ddagger$ &       \cellcolor{red!12}$0.07^\star$ &                                   -- &                                  -- &                                  -- &                                 -- &                                   -- \\
              \textit{Personal concerns} &                                      &                                      &                                      &                                      &                                      &                                     &                                     &                                    &                                      \\
                  \hspace{3pt}Relativity &     \cellcolor{red!27}$0.05^\dagger$ &     \cellcolor{red!27}$0.06^\dagger$ &                                   -- &                                   -- &                                   -- &                                  -- &                                  -- &                                 -- &                                   -- \\
                        \hspace{3pt}Time &    \cellcolor{red!45}$0.06^\ddagger$ &                                   -- &       \cellcolor{red!12}$0.06^\star$ &                                   -- &                                   -- &                                  -- &                                  -- &                                 -- &                                   -- \\
                        \hspace{3pt}Work &                                   -- &       \cellcolor{red!12}$0.06^\star$ &                                   -- &                                   -- &                                   -- &                                  -- &                                  -- &                                 -- &                                   -- \\
                     \hspace{3pt}Leisure &  \cellcolor{blue!45}$-0.07^\ddagger$ &  \cellcolor{blue!45}$-0.07^\ddagger$ &  \cellcolor{blue!45}$-0.07^\ddagger$ &  \cellcolor{blue!45}$-0.09^\ddagger$ &  \cellcolor{blue!45}$-0.12^\ddagger$ &  \cellcolor{blue!27}$-0.09^\dagger$ &                                  -- &                                 -- &                                   -- \\
                       \hspace{3pt}Money &  \cellcolor{blue!45}$-0.06^\ddagger$ &                                   -- &  \cellcolor{blue!45}$-0.05^\ddagger$ &  \cellcolor{blue!45}$-0.06^\ddagger$ &                                   -- &  \cellcolor{blue!45}$-0.1^\ddagger$ &                                  -- &                                 -- &  \cellcolor{blue!45}$-0.12^\ddagger$ \\
           \hspace{3pt}Informal language &  \cellcolor{blue!45}$-0.07^\ddagger$ &  \cellcolor{blue!45}$-0.07^\ddagger$ &  \cellcolor{blue!45}$-0.06^\ddagger$ &                                   -- &  \cellcolor{blue!45}$-0.12^\ddagger$ &                                  -- &    \cellcolor{blue!12}$-0.09^\star$ &                                 -- &                                   -- \\
                    \hspace{3pt}Netspeak &  \cellcolor{blue!45}$-0.07^\ddagger$ &  \cellcolor{blue!45}$-0.06^\ddagger$ &  \cellcolor{blue!45}$-0.06^\ddagger$ &     \cellcolor{blue!12}$-0.06^\star$ &   \cellcolor{blue!45}$-0.1^\ddagger$ &    \cellcolor{blue!12}$-0.08^\star$ &                                  -- &                                 -- &  \cellcolor{blue!45}$-0.15^\ddagger$ \\
                      \hspace{3pt}Assent &                                   -- &                                   -- &                                   -- &                                   -- &     \cellcolor{blue!12}$-0.06^\star$ &                                  -- &                                  -- &                                 -- &                                   -- \\
  \bottomrule
  \end{tabular}

  \caption{Differences between the linguistic and psycholinguistic variables obtained from LIWC between users treatment groups and control users. Statistical significance is based on t-test with Bonferroni adjustment. Numbers in the cells are effect sizes (Cohen's $d$). $\star$, $\dagger$, and $\ddagger$ respectively show adjusted p-values of ${<}0.01$, ${<}0.001$ and ${<}0.0001$. This is also indicated by shading colors, red shading shows that the corresponding category is observed with more intensity in the treatment group, where as blue shading shows more significance in the control group. For brevity, only significant attributes are shown.}
  \label{tab:liwc-all}

  \end{table}

In general, we observe a variety of differences in language use between the diagnosed and the control groups. The effect sizes range from a very small effect to medium ($0.05{<}d{<}0.5$), which is typical of large datasets \cite{sakai2016statistical}. These differences span across both the linguistic style and psychological attributes. Particularly, we observe that clout, an attribute regarding language that indicates high social status, is more prominent among control users compared to the depression and anxiety groups. Related work in psychology shows that depression and anxiety are more prevalent in lower socioeconomic status demographics \cite{murphy1991depression}, which is in line with the observation in our dataset. Next, authentic language is stronger among most of the mental health groups compared with control groups.  This correlates with the use of personal and first person pronouns. This is backed by prior research showing that fewer self-references are associated with unauthentic language \cite{newman2003lying}. We also observe that among most of the conditions the usage of first-person singular pronouns is higher than in control groups. Related studies arrive at the same conclusion that people with mental health conditions tend to use more first-person pronouns and references, possibly because mental health conditions result in greater self-focused attention and rumination \cite{bucci1981language,pennebaker2015development,watkins2002rumination,van2014web}.

Social processes, the language related to human interaction and sharing, appears to be higher in depressed, anxiety and PTSD users. A potential explanation for this is that sharing and support-seeking behaviors have been shown to improve an individual's state of mind \cite{turner1983social,de2017language}.
People with conditions such as bipolar, depression and anxiety show significantly more female references. This might point to a gender bias in the corpus towards males, which would shift an increased preoccupation with romantic relationships towards references of female partners. Cognitive processes (markers of cognitive activity associated with rational thought and argumentation) are also observed more in people with mental health conditions. This could be due to the observation that these groups tend to express their feelings more. Other attributes such as health, focus on past time, and biological processes are observed more in conditions such as depression. Unsurprisingly, attributes like leisure and money are more prevalent among controls.

\section{Experiments}
\label{sec:method}

\begin{table}[t]
{\scriptsize
\renewcommand\arraystretch{1.2}
\renewcommand\tabcolsep{6pt}
\begin{tabularx}{\textwidth}{ X c c c c c c c c c c }
\toprule
 & \textbf{Depression} & \textbf{ADHD}  & \textbf{Anxiety}  & \textbf{Bipolar}  & \textbf{PTSD}  & \textbf{Autism}  & \textbf{OCD}  & \textbf{Schizophrenia}  & \textbf{Eating} & \textbf{Multi-label} \\
\midrule
\multirow{3}{*}{\shortstack[l]{\textbf{Logistic} \\ \textbf{Regression}\\ BoW features}}  & \textbf{P=85.00} & \textbf{P=88.60}  & \textbf{P=85.80}  & \textbf{P=87.06}  & P=92.44  & \textbf{P=93.18}  & \textbf{P=95.00}  & \textbf{P=100.00}  & P=0.00 & P=17.22 \\
 & R=28.65 & R=19.22  & R=27.04  & R=21.57  & R=19.71  & R=7.93  & R=4.87  & R=2.25  & R=0.00 & R=24.38 \\
 & F=42.85 & F=31.59  & F=41.13  & F=34.58  & F=32.50  & F=14.62  & F=9.27  & F=4.40  & F=0.00 & F=19.72 \\
\midrule
\multirow{3}{*}{\shortstack[l]{\textbf{XGBoost}\\ BoW features}}& P=81.40 & P=85.71  & P=83.52  & P=84.94  & P=89.83  & P=84.09  & P=86.96  & P=86.84  & P=94.12 & P=16.87 \\
 & R=29.49 & R=21.25  & R=31.46  & R=30.31  & R=37.99  & R=21.47  & R=25.64  & R=24.72  & R=14.29 & R=29.38 \\
 & F=43.29 & F=34.05  & F=45.71  & F=44.68  & F=53.40  & F=34.21  & F=39.60  & F=38.48  & F=24.81 & F=20.51 \\
\midrule
\multirow{3}{*}{\shortstack[l]{\textbf{Linear SVM}\\ BoW features}} & P=78.70 & P=83.74  & P=81.76  & P=84.36  &\bf P=93.12  & P=91.19  & P=86.29  & P=93.06  & \textbf{P=100.00} & P=19.20 \\
 & R=39.75 & R=30.69  & R=39.88  & R=37.21  & R=41.22  & R=28.05  & R=27.44  & R=25.09  & R=13.39 & R=34.86 \\
 & F=52.82 & F=44.92  & F=53.61  & \bf F=51.64  & F=57.14  & F=42.90  & F=41.63  & F=39.53  & F=23.62 & F=23.71 \\
\midrule
\multirow{3}{*}{\shortstack[l]{\textbf{Supervised} \\ \textbf{FastText}}}& P=66.80 & P=62.11  & P=67.63  & P=62.63  & P=70.76  & P=68.47  & P=65.02  & P=69.23  & P=64.15 & \bf P=23.02 \\
 & R=44.70 &\bf R=37.77  & R=44.54  &\bf R=42.74  & R=48.57  & R=39.07  & R=33.85  & R=33.71  & R=30.36 & R=44.44 \\
 &\bf F=53.56 &\bf F=46.98  &\bf F=53.71  & F=50.81  &\bf F=57.60  &\bf F=49.75  &\bf F=44.52  &\bf F=45.34  & F=41.21 & \textbf{F=27.83} \\
\midrule
\multirow{3}{*}{\shortstack[l]{\textbf{Convolutional}\\ \textbf{Neural} \\ \textbf{Network (CNN)}}}& P=45.85 & P=37.86 & P=51.33 & P=37.21& P=47.00& P=30.82& P=22.97& P=30.12& P=38.56 & P=20.65 \\
&\bf R=56.07 & R=32.77 &\bf R=45.01 & R=39.53 &\bf R=65.95 &\bf R=50.68 &\bf R=57.95 &\bf R=47.94 &\bf R=52.68  & \bf R=48.12  \\
& F=50.45 & F=35.13 & F=47.96 & F=38.34 & F=54.88 & F=38.33 & F=32.90 & F=36.99 &\bf F=44.53  & F=26.55  \\
\bottomrule
\end{tabularx}
}
\caption{Classification performance on target users in \datasetname.
All methods rely on the official \datasetname split introduced in Section \ref{sec:data} for training, tuning, and testing.
We report the performance of each classifier as binary predictor (i.e., given a condition, predict if a user belongs to the diagnosed or control group) as well as  in a multi-label, multi-class setting (i.e., given a user, predict if and which conditions they were diagnosed with).
}
\label{table:baseline_classifiers}
\end{table}

In order to provide a more thorough understanding of \datasetname, we explored several classification methods for each of the conditions included in the dataset. We train our classifiers both in a binary and multi-label multi-class setting.
For the binary tasks, all classifiers are trained only on the subset of diagnosed users in \datasetname who are associated with any condition and their respective control users.
Recall that, as discussed in \ref{sec:stats}, users can be diagnosed with one or more conditions; as such, we also evaluate the baseline classifiers in a multi-class, multi-label setting.
In other words, given a user, classifiers are trained to predict if and which conditions the user was diagnosed with.
In detail, we consider the following methods:

\textbf{Logistic regression}: we trained a simple discriminative model using bag-of-words features extracted over all posts by all users in the training set.
Posts were concatenated and split on non-alphanumeric characters; we normalized tokens by folding them to lowercase.
Words occurring less than 20 times in the training set were removed. Features were weighted using \textit{tf-idf} and $\ell_2$-normalized.

\textbf{XGBoost} \cite{chen2016xgboost}: we evaluated the performance of an ensemble of decision trees. For this model, the same features described above were used. We set the number of estimators to $100$.

\textbf{Support vector machine (SVM)}: we included a SVM classifier with linear kernel among our baselines. Like the methods above, this model was trained on \textit{tf-idf} bag-of-words features.

\textbf{Supervised FastText} \cite{joulin2016bag}: we trained a shallow neural net model using FastText. After tuning on the development set, we set the dimension of the hidden layer to 100. The model leverages sub-word information by using character ngrams of size 3 to 6; it was trained for 100 epochs.

\textbf{Convolutional neural network (CNN)}: Finally, we tested a CNN approach to learn ngram sequences indicative of each condition. We used a model architecture based on the work of~\citet{Kim2014ConvolutionalNN}, with a filter size of 3 and a pooling size of 4. A model was trained for each condition, and each model was trained for 100 epochs. Each user's posts are concatenated and truncated to 15,000 tokens. This representation combines tractability with density. We do not pad posts individually hence no space is wasted by including short posts. Each token is represented by the FastText embeddings from above.

\medskip
Results of the classification methods are reported in Table~\ref{table:baseline_classifiers}. All classifiers are trained, tuned, and tested on binary labels, except for the results reported in the rightmost column (``multi-label'').
Overall, FastText obtained the best performance in terms of F1 score across all conditions except bipolar, in which it is outperformed by the SVM model, and eating disorder, where it is outperformed by a CNN.
FastText also outperforms all other classifiers in the multi-label, multi-class test setting.
We observe that models trained on bag-of-words features favor precision over recall, while the two neural models we tested offer more balanced performance.
This can be explained by the fact that bag-of-words models are more likely to give more weight to strongly indicative text features, potentially causing some overfitting on the training data.

Comparing across conditions, we note that the performance on models is strongly affected by the number of diagnosed users in the training set.
All classifiers struggle on the three conditions with fewest users (OCD, schizophrenia, and eating). This observation is consistent with the analysis on LIWC categories reported in Section~\ref{sec:analysis}.

Finally, we note that the simpler neural model (FastText) outperformed the richer CNN method.
We explain this result by observing that FastText considers all posts by a user, while CNN samples a portion of them.
This strategy, which is motivated by efficiency reasons, was based on the work of~\citet{yates2017depression}, in which sampling was found to be an effective strategy to make the classification problem tractable.
However, this result suggests that novel methods ought to be studied to achieve a robust sampling strategy across all conditions included in \datasetname.


\section{Conclusion}
We presented \datasetname, a large dataset of Reddit users with diverse mental health conditions and matched control users. Our dataset was constructed using high-precision diagnosis patterns and carefully selected control users. To our knowledge, \datasetname is the largest dataset that supports a variety of mental health conditions and is up to two orders of magnitude larger than the largest published similar resource. We further examined differences in language use between mental health condition (diagnosed users) and control groups as measured by various linguistic and psychological signals. Several text classification methods were explored to identify the diagnosed users, with FastText being the most effective approach overall. We make our dataset available to the community, and hope that it will foster further research into these problems and enhance reproducibility of suggested approaches.

\bibliography{references.bib}

\begin{thebibliography}{}

\bibitem[\protect\citename{Althoff \bgroup et al.\egroup
  }2016]{Althoff2016LargescaleAO}
Tim Althoff, Kevin Clark, and Jure Leskovec.
\newblock 2016.
\newblock Large-scale analysis of counseling conversations: An application of
  natural language processing to mental health.
\newblock In {\em TACL}.

\bibitem[\protect\citename{{American Psychiatric
  Association}}2013]{apa2013diagnostic}
{American Psychiatric Association}.
\newblock 2013.
\newblock {\em Diagnostic and Statistical Manual of Mental Disorders
  ({DSM-5})}.
\newblock American Psychiatric Association Publishing.

\bibitem[\protect\citename{Benton \bgroup et al.\egroup
  }2017]{Benton2017MultitaskLF}
Adrian Benton, Margaret Mitchell, and Dirk Hovy.
\newblock 2017.
\newblock Multitask learning for mental health conditions with limited social
  media data.
\newblock In {\em EACL}.

\bibitem[\protect\citename{Bucci and Freedman}1981]{bucci1981language}
Wilma Bucci and Norbert Freedman.
\newblock 1981.
\newblock The language of depression.
\newblock {\em Bulletin of the Menninger Clinic}, 45(4):334.

\bibitem[\protect\citename{Carmel \bgroup et al.\egroup
  }2002]{carmel2002automatic}
David Carmel, Eitan Farchi, Yael Petruschka, and Aya Soffer.
\newblock 2002.
\newblock Automatic query refinement using lexical affinities with maximal
  information gain.
\newblock In {\em Proceedings of the 25th annual international ACM SIGIR
  conference on Research and development in information retrieval}, pages
  283--290. ACM.

\bibitem[\protect\citename{Chen and Guestrin}2016]{chen2016xgboost}
Tianqi Chen and Carlos Guestrin.
\newblock 2016.
\newblock {XGBoost}: A scalable tree boosting system.
\newblock In {\em Proceedings of the 22nd acm sigkdd international conference
  on knowledge discovery and data mining}, pages 785--794. ACM.

\bibitem[\protect\citename{Cohan \bgroup et al.\egroup
  }2017]{cohan2017triaging}
Arman Cohan, Sydney Young, Andrew Yates, and Nazli Goharian.
\newblock 2017.
\newblock Triaging content severity in online mental health forums.
\newblock {\em Journal of the Association for Information Science and
  Technology (JASIST)}.

\bibitem[\protect\citename{Cohen}1988]{cohen1988statistical}
Jacob Cohen.
\newblock 1988.
\newblock Statistical power analysis for the behavioral sciences 2nd edition.

\bibitem[\protect\citename{Coppersmith \bgroup et al.\egroup
  }2014a]{coppersmith2014quantifying}
Glen Coppersmith, Mark Dredze, and Craig Harman.
\newblock 2014a.
\newblock Quantifying mental health signals in {Twitter}.
\newblock In {\em Proceedings of the Workshop on Computational Linguistics and
  Clinical Psychology: From Linguistic Signal to Clinical Reality}, pages
  51--60.

\bibitem[\protect\citename{Coppersmith \bgroup et al.\egroup
  }2014b]{Coppersmith2014MeasuringPT}
Glen Coppersmith, Craig Harman, and Mark Dredze.
\newblock 2014b.
\newblock Measuring post traumatic stress disorder in {Twitter}.
\newblock In {\em ICWSM}.

\bibitem[\protect\citename{Coppersmith \bgroup et al.\egroup
  }2015a]{coppersmith2015adhd}
Glen Coppersmith, Mark Dredze, Craig Harman, and Kristy Hollingshead.
\newblock 2015a.
\newblock From {ADHD} to {SAD}: Analyzing the language of mental health on
  {Twitter} through self-reported diagnoses.
\newblock In {\em Proceedings of the 2nd Workshop on Computational Linguistics
  and Clinical Psychology: From Linguistic Signal to Clinical Reality}, pages
  1--10.

\bibitem[\protect\citename{Coppersmith \bgroup et al.\egroup
  }2015b]{Coppersmith2015CLPsych2S}
Glen Coppersmith, Mark Dredze, Craig Harman, Kristy Hollingshead, and Margaret
  Mitchell.
\newblock 2015b.
\newblock {CLPsych} 2015 shared task: Depression and {PTSD} on {Twitter}.
\newblock In {\em CLPsych HLT-NAACL}.

\bibitem[\protect\citename{Coppersmith \bgroup et al.\egroup
  }2017]{coppersmith2017scalable}
Glen Coppersmith, Casey Hilland, Ophir Frieder, and Ryan Leary.
\newblock 2017.
\newblock Scalable mental health analysis in the clinical whitespace via
  natural language processing.
\newblock In {\em Biomedical \& Health Informatics (BHI), 2017 IEEE EMBS
  International Conference on}, pages 393--396. IEEE.

\bibitem[\protect\citename{De~Choudhury and K{\i}c{\i}man}2017]{de2017language}
Munmun De~Choudhury and Emre K{\i}c{\i}man.
\newblock 2017.
\newblock The language of social support in social media and its effect on
  suicidal ideation risk.
\newblock In {\em Proceedings of the... International AAAI Conference on
  Weblogs and Social Media. International AAAI Conference on Weblogs and Social
  Media}, volume 2017, page~32. NIH Public Access.

\bibitem[\protect\citename{De~Choudhury \bgroup et al.\egroup
  }2013a]{de2013major}
Munmun De~Choudhury, Scott Counts, and Eric Horvitz.
\newblock 2013a.
\newblock Major life changes and behavioral markers in social media: case of
  childbirth.
\newblock In {\em Proceedings of the 2013 conference on Computer supported
  cooperative work}, pages 1431--1442. ACM.

\bibitem[\protect\citename{De~Choudhury \bgroup et al.\egroup
  }2013b]{Choudhury2013PredictingDV}
Munmun De~Choudhury, Michael Gamon, Scott Counts, and Eric Horvitz.
\newblock 2013b.
\newblock Predicting depression via social media.
\newblock In {\em ICWSM}.

\bibitem[\protect\citename{Desmet and Hoste}2018]{desmet2018suicide}
Bart Desmet and Veronique Hoste.
\newblock 2018.
\newblock Online suicide prevention through optimised text classification.
\newblock {\em Information Sciences}, 439-440:61--78.

\bibitem[\protect\citename{Ernala \bgroup et al.\egroup
  }2017]{Ernala2017LinguisticMI}
Sindhu~Kiranmai Ernala, Asra~F. Rizvi, Michael~L. Birnbaum, John~M. Kane, and
  Munmun~De Choudhury.
\newblock 2017.
\newblock Linguistic markers indicating therapeutic outcomes of social media
  disclosures of schizophrenia.
\newblock {\em PACMHCI}, 1:43:1--43:27.

\bibitem[\protect\citename{Hovy and Spruit}2016]{hovy2016social}
Dirk Hovy and Shannon~L Spruit.
\newblock 2016.
\newblock The social impact of natural language processing.
\newblock In {\em Proceedings of the 54th Annual Meeting of the Association for
  Computational Linguistics (Volume 2: Short Papers)}, volume~2, pages
  591--598.

\bibitem[\protect\citename{Joulin \bgroup et al.\egroup }2016]{joulin2016bag}
Armand Joulin, Edouard Grave, Piotr Bojanowski, and Tomas Mikolov.
\newblock 2016.
\newblock Bag of tricks for efficient text classification.
\newblock {\em arXiv preprint arXiv:1607.01759}.

\bibitem[\protect\citename{Kim}2014]{Kim2014ConvolutionalNN}
Yoon Kim.
\newblock 2014.
\newblock Convolutional neural networks for sentence classification.
\newblock In {\em EMNLP}.

\bibitem[\protect\citename{Kshirsagar \bgroup et al.\egroup
  }2017]{kshirsagar2017detecting}
Rohan Kshirsagar, Robert Morris, and Sam Bowman.
\newblock 2017.
\newblock Detecting and explaining crisis.
\newblock {\em arXiv preprint arXiv:1705.09585}.

\bibitem[\protect\citename{Losada and Crestani}2016]{Losada2016ATC}
David~E. Losada and Fabio Crestani.
\newblock 2016.
\newblock A test collection for research on depression and language use.
\newblock In {\em CLEF}.

\bibitem[\protect\citename{MacAvaney \bgroup et al.\egroup
  }2018]{macavaney2018rsdd}
Sean MacAvaney, Bart Desmet, Arman Cohan, Luca Soldaini, Andrew Yates, Ayah
  Zirikly, and Nazli Goharian.
\newblock 2018.
\newblock {RSDD-Time}: Temporal annotation of self-reported mental health
  diagnoses.
\newblock In {\em Proceedings of the Fifth Workshop on Computational
  Linguistics and Clinical Psychology: From Keyboard to Clinic}, pages
  168--173.

\bibitem[\protect\citename{Milne \bgroup et al.\egroup
  }2016]{Milne2016CLPsych2S}
David~N. Milne, Glen Pink, Ben Hachey, and Rafael~Alejandro Calvo.
\newblock 2016.
\newblock {CLPsych} 2016 shared task: Triaging content in online peer-support
  forums.
\newblock In {\em CLPsych@HLT-NAACL}.

\bibitem[\protect\citename{Mitchell \bgroup et al.\egroup
  }2015]{mitchell2015quantifying}
Margaret Mitchell, Kristy Hollingshead, and Glen Coppersmith.
\newblock 2015.
\newblock Quantifying the language of schizophrenia in social media.
\newblock In {\em Proceedings of the 2nd workshop on Computational linguistics
  and clinical psychology: From linguistic signal to clinical reality}, pages
  11--20.

\bibitem[\protect\citename{Mowery \bgroup et al.\egroup
  }2016]{mowery2016towards}
Danielle~L Mowery, Albert Park, Craig Bryan, and Mike Conway.
\newblock 2016.
\newblock Towards automatically classifying depressive symptoms from {Twitter}
  data for population health.
\newblock In {\em Proceedings of the Workshop on Computational Modeling of
  People's Opinions, Personality, and Emotions in Social Media (PEOPLES)},
  pages 182--191.

\bibitem[\protect\citename{Mowery \bgroup et al.\egroup
  }2017a]{mowery2017feature}
Danielle Mowery, Craig Bryan, and Mike Conway.
\newblock 2017a.
\newblock Feature studies to inform the classification of depressive symptoms
  from {Twitter} data for population health.
\newblock {\em arXiv preprint arXiv:1701.08229}.

\bibitem[\protect\citename{Mowery \bgroup et al.\egroup
  }2017b]{mowery2017understanding}
Danielle Mowery, Hilary Smith, Tyler Cheney, Greg Stoddard, Glen Coppersmith,
  Craig Bryan, and Mike Conway.
\newblock 2017b.
\newblock Understanding depressive symptoms and psychosocial stressors on
  {Twitter}: a corpus-based study.
\newblock {\em Journal of medical Internet research}, 19(2).

\bibitem[\protect\citename{Murphy \bgroup et al.\egroup
  }1991]{murphy1991depression}
Jane~M Murphy, Donald~C Olivier, Richard~R Monson, Arthur~M Sobol, Elizabeth~B
  Federman, and Alexander~H Leighton.
\newblock 1991.
\newblock Depression and anxiety in relation to social status: A prospective
  epidemiologic study.
\newblock {\em Archives of General Psychiatry}, 48(3):223--229.

\bibitem[\protect\citename{Newman \bgroup et al.\egroup }2003]{newman2003lying}
Matthew~L Newman, James~W Pennebaker, Diane~S Berry, and Jane~M Richards.
\newblock 2003.
\newblock Lying words: Predicting deception from linguistic styles.
\newblock {\em Personality and social psychology bulletin}, 29(5):665--675.

\bibitem[\protect\citename{Pavalanathan and
  De~Choudhury}2015]{Pavalanathan:2015:IMM:2740908.2743049}
Umashanthi Pavalanathan and Munmun De~Choudhury.
\newblock 2015.
\newblock Identity management and mental health discourse in social media.
\newblock In {\em Proceedings of the 24th International Conference on World
  Wide Web}, WWW '15 Companion, pages 315--321, New York, NY, USA. ACM.

\bibitem[\protect\citename{Pennebaker \bgroup et al.\egroup
  }2015]{pennebaker2015development}
James~W Pennebaker, Ryan~L Boyd, Kayla Jordan, and Kate Blackburn.
\newblock 2015.
\newblock The development and psychometric properties of {LIWC2015}.
\newblock Technical report.

\bibitem[\protect\citename{Preotiuc-Pietro \bgroup et al.\egroup
  }2015]{PreotiucPietro2015TheRO}
Daniel Preotiuc-Pietro, Johannes~C. Eichstaedt, Gregory~J. Park, Maarten Sap,
  Laura Smith, Victoria Tobolsky, H.~Andrew Schwartz, and Lyle~H. Ungar.
\newblock 2015.
\newblock The role of personality, age, and gender in tweeting about mental
  illness.
\newblock In {\em CLPsych HLT-NAACL}.

\bibitem[\protect\citename{Resnik \bgroup et al.\egroup
  }2013]{Resnik2013UsingTM}
Philip Resnik, Anderson Garron, and Rebecca Resnik.
\newblock 2013.
\newblock Using topic modeling to improve prediction of neuroticism and
  depression in college students.
\newblock In {\em EMNLP}.

\bibitem[\protect\citename{Resnik \bgroup et al.\egroup }2015]{Resnik2015TheUO}
Philip Resnik, William Armstrong, Leonardo Claudino, and Thang Nguyen.
\newblock 2015.
\newblock The {University of Maryland} {CLPsych} 2015 shared task system.
\newblock In {\em CLPsych HLT-NAACL}.

\bibitem[\protect\citename{Sakai}2016]{sakai2016statistical}
Tetsuya Sakai.
\newblock 2016.
\newblock Statistical significance, power, and sample sizes: A systematic
  review of {SIGIR} and {TOIS}, 2006-2015.
\newblock In {\em Proceedings of the 39th International ACM SIGIR conference on
  Research and Development in Information Retrieval}, pages 5--14. ACM.

\bibitem[\protect\citename{Soldaini and Yom-Tov}2017]{soldaini2017inferring}
Luca Soldaini and Elad Yom-Tov.
\newblock 2017.
\newblock Inferring individual attributes from search engine queries and
  auxiliary information.
\newblock In {\em Proceedings of the 26th International Conference on World
  Wide Web (WWW '17)}, pages 293--301.

\bibitem[\protect\citename{Strine \bgroup et al.\egroup
  }2008]{strine2008depression}
Tara~W Strine, Ali~H Mokdad, Lina~S Balluz, Olinda Gonzalez, Raquel Crider,
  Joyce~T Berry, and Kurt Kroenke.
\newblock 2008.
\newblock Depression and anxiety in the {United States}: findings from the 2006
  behavioral risk factor surveillance system.
\newblock {\em Psychiatric services}, 59(12):1383--1390.

\bibitem[\protect\citename{{\v{S}}uster \bgroup et al.\egroup
  }2017]{vsuster2017short}
Simon {\v{S}}uster, St{\'e}phan Tulkens, and Walter Daelemans.
\newblock 2017.
\newblock A short review of ethical challenges in clinical natural language
  processing.
\newblock {\em arXiv preprint arXiv:1703.10090}.

\bibitem[\protect\citename{Tsugawa \bgroup et al.\egroup
  }2015]{Tsugawa2015RecognizingDF}
Sho Tsugawa, Yusuke Kikuchi, Fumio Kishino, Kosuke Nakajima, Yuichi Itoh, and
  Hiroyuki Ohsaki.
\newblock 2015.
\newblock Recognizing depression from {Twitter} activity.
\newblock In {\em CHI}.

\bibitem[\protect\citename{Turner \bgroup et al.\egroup
  }1983]{turner1983social}
R~Jay Turner, B~Gail Frankel, and Deborah~M Levin.
\newblock 1983.
\newblock Social support: Conceptualization, measurement, and implications for
  mental health.
\newblock {\em Research in community \& mental health}.

\bibitem[\protect\citename{Van~der Zanden \bgroup et al.\egroup
  }2014]{van2014web}
Rianne Van~der Zanden, Keshia Curie, Monique Van~Londen, Jeannet Kramer, Gerard
  Steen, and Pim Cuijpers.
\newblock 2014.
\newblock Web-based depression treatment: Associations of clients׳ word use
  with adherence and outcome.
\newblock {\em Journal of Affective Disorders}, 160:10--13.

\bibitem[\protect\citename{Watkins and Brown}2002]{watkins2002rumination}
E~Watkins and RG~Brown.
\newblock 2002.
\newblock Rumination and executive function in depression: An experimental
  study.
\newblock {\em Journal of Neurology, Neurosurgery \& Psychiatry},
  72(3):400--402.

\bibitem[\protect\citename{Welch}1947]{welch1947generalization}
Bernard~L Welch.
\newblock 1947.
\newblock The generalization ofstudent's' problem when several different
  population variances are involved.
\newblock {\em Biometrika}, 34(1/2):28--35.

\bibitem[\protect\citename{Yates and Goharian}2013]{yates2013adrtrace}
Andrew Yates and Nazli Goharian.
\newblock 2013.
\newblock {ADRTrace}: detecting expected and unexpected adverse drug reactions
  from user reviews on social media sites.
\newblock In {\em European Conference on Information Retrieval}, pages
  816--819. Springer.

\bibitem[\protect\citename{Yates \bgroup et al.\egroup
  }2017]{yates2017depression}
Andrew Yates, Arman Cohan, and Nazli Goharian.
\newblock 2017.
\newblock Depression and self-harm risk assessment in online forums.
\newblock In {\em Proceedings of the 2017 Conference on Empirical Methods in
  Natural Language Processing}, pages 2968--2978, Copenhagen, Denmark,
  September. Association for Computational Linguistics.

\bibitem[\protect\citename{Yom-Tov and Gabrilovich}2013]{yom2013postmarket}
Elad Yom-Tov and Evgeniy Gabrilovich.
\newblock 2013.
\newblock Postmarket drug surveillance without trial costs: discovery of
  adverse drug reactions through large-scale analysis of web search queries.
\newblock {\em Journal of medical Internet research}, 15(6).

\end{thebibliography}
\bibliographystyle{acl}

\end{document}